\documentclass[letterpaper, 10 pt, conference]{ieeeconf}  % Comment this line out if you need a4paper

\IEEEoverridecommandlockouts                              % This command is only needed if 
\usepackage{graphicx}

\usepackage{latexsym}
\usepackage{color}
\usepackage{supertabular}
\usepackage{multirow}
\usepackage{setspace}
\usepackage{amsmath}
\usepackage{amsfonts}

\title{\LARGE \bf
The Fractal Hand-II: Reviving a Classic Mechanism for Contemporary Grasping Challenges
}
\author{Malcolm Tisdale and Joel W. Burdick% <-this % stops a space
\thanks{*This work was supported in part by the Caltech Center for Autonomous Systems and Technologies, and the Robert I. and Winifred E. Gardner SURF Fellowship}% <-this % stops a space
\thanks{Authors are with the Depart. of Mechanical and Civil Engineering,
        California Institute of Technology, Pasadena, CA 91125, USA
        {\tt\small [mtisdale,jburdick]@caltech.edu}}%
}

\begin{document}

\maketitle
\thispagestyle{empty}
\pagestyle{empty}

%%%%%%%%%%%%%%%%%%%%%%%%%%%%%%%%%%%%%%%%%%%%%%%%%%%%%%%%%%%%%%%%%%%%%%%%%%%%%%%%
\begin{abstract}

This paper, and its companion \cite{Companion_Paper}, propose a new {\em fractal} robotic gripper, drawing inspiration from the century-old {\em Fractal Vise}. The unusual synergistic properties allow it to passively conform to diverse objects using only one actuator. Designed to be easily integrated with prevailing parallel jaw grippers, it alleviates the complexities tied to perception and grasp planning, especially when dealing with unpredictable object poses and geometries. We build on the foundational principles of the Fractal Vise to a broader class of gripping mechanisms, and also address the limitations that had led to its obscurity. Two {\em Fractal Fingers}, coupled by a closing actuator, can form an adaptive and synergistic {\em Fractal Hand}. We articulate a design methodology for low cost, easy to fabricate, large workspace, and compliant Fractal Fingers. The companion paper delves into the kinematics and grasping properties of a specific class of Fractal Fingers and Hands. 

% joels below
%This paper introduces a novel {\em Fractal Hand} robotic gripper. The hand has only 1 actuator, but $(2^{n+1}-1)$ joints, where $n$ is a design parameter that defines the depth of the fingers' tree structures. The hand is {\em synergistic} in its operation (because its joint movements are highly coupled through the hand's interaction with the grasped object), but it is not anthropomorphic.  The basic finger and hand geometry, governing kinematics, and quasi-statics mechanics of {\em rigid} and {\em compliant} versions of the hand are developed.  These analyses remarkably show that under very mild constraints on the hand design, the hand is compliantly stable at every equilibrium condition.  Thefore, the Fractal Hand adapts to a wide range of planar objects with a single design.  A companion paper \cite{Companion_Paper} introduces a design methodology for this new class of robot hands, and multiple prototypes. 
\end{abstract}

% Robust, yet fast, grasp planning for adversarial objects remains an unresolved challenge in robotics. Synergistic grippers tackle this problem by trading computational with mechanical complexity, thus leveraging environmental features to achieve successful grasps. The expired patent US1059545A describes a highly conformable universal vise, also known as a fractal vise, whose design principles inspire our approach. We have developed a synergistic, nonanthropomorphic, \(Fractal\) \(Hand\) that is agnostic to an objects' shape and pose. The \(Fractal\) \(Hand\) enables grasping of adversarial objects with minimal grasp planning. This paper presents the design methodology and experimentation of a  fractal hand embodiment, as applied to robotic manipulation. We first develop a type and dimensional synthesis method that quantifies the scale variant complexity — the range and variation of detail at different scales — of objects in a data set to generate a customised Fractal Hand embodiment. We then show the hand grasping adversarial objects using a fast, simple, grasp planning pipeline. This work aims to explore and highlight the potential of the extensive, yet largely unexplored, design space of fractal hands.

%%%%%%%%%%%%%%%%%%%%%%%%%%%%%%%%%%%%%%%%%%%%%%%%%%%%%%%%%%%%%%%%%%%%%%%%%%%%%%%%
\section{INTRODUCTION}

The desire to rapidly plan robust grasps for complex objects presents an ongoing challenge in the field of robotics. Various analytical and empirical strategies have been employed to address this problem. Analytic methods synthesize optimal grasps using grasp quality measures based on known object geometry and mechanics \cite{ferrariCanny, GraspQualityMeasures, CharacterisingGraspMetrics}, but can be undermined by mechanical and perceptual uncertainties \cite{Kappler, metricflaws}. 

%Empirical approaches are motivated by the practically important discrepancy between real-world experience and analytic approaches. Deep neural networks, paired with high fidelity perception, have been used to achieve capable grasps by incorporating perception uncertainties into the grasping pipeline. This necessitates high fidelity perception to minimize errors. Deep learning has led to end-to-end grasp synthesis using very large training sets generated in simulation or experiments \cite{dexnet2, newbury2023deep}. However, both analytical and empirical approaches still suffer from physical and environmental uncertainties, all while having considerable computational cost. 

Empirical approaches are motivated by the practically important discrepancy between real-world experience and analytic approaches. Deep neural networks, paired with high fidelity perception, have been used to achieve capable grasps by incorporating perception uncertainties into the grasping pipeline \cite{dexnet2, newbury2023deep}. However, this approach has considerable sensory and computation cost, still falling short of real-time-speed functionality. 

\begin{figure}[t]
    \centering
    \includegraphics[scale=0.35]{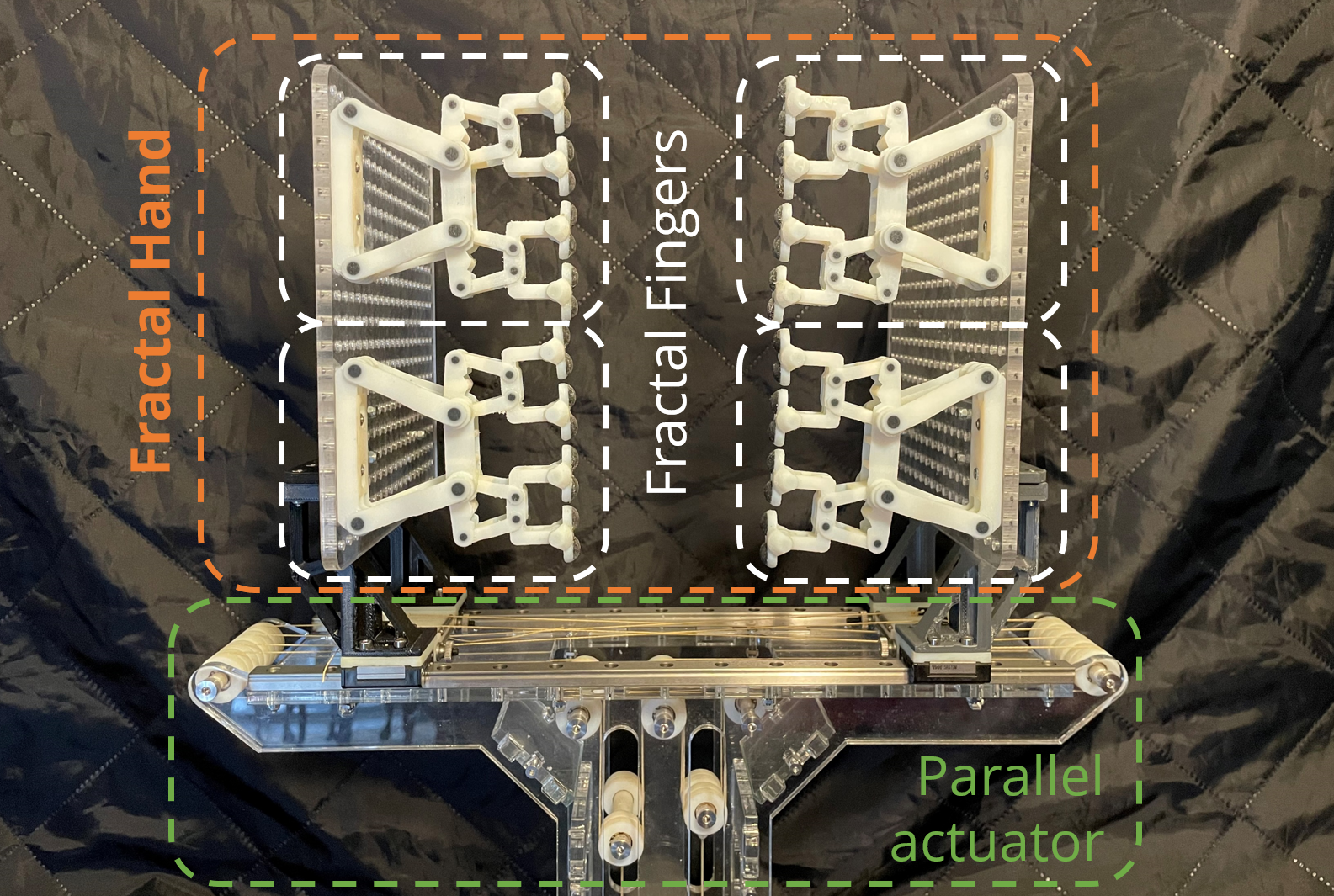}
    \caption{Picture of a 4-fingered planar Fractal Hand prototype mounted on a tendon-driven, passive center, parallel hand closing actuator.}
    \label{fig:handonwrist}
    \vskip -0.2 true in
\end{figure}

To address these challenges, Bicchi and others have explored an alternative paradigm—an approach that posits intelligence as the product of interaction between a system and its environment, coined \textit{embodied intelligence} \cite{embodiedintelligence, Mengaldo2022}. Bicchi, Dollar, Cutkosky, and others have shown that reaction wrenches on a gripper, generated by contact forces from the environment, can be used to deform the grippers and achieve successful grasps with under-actuated systems \cite{graspingwithsofthands,sdmhand,oceanonehands}. This has led to the development of {\em synergistic} robotic hands, which use specific joint couplings to reduce the hands' active Degrees of Freedom (DoF) while maintaining the capability for various types of grasping. This synergistic approach to gripper design trades the computational intensity of the grasp planning problem with mechanical design.  However, these synergistic hand designs may be limited to a given set of grasping tasks.

Virtually all synergistic hand designs are anthropomorphic, having multiple serial chain finger mechanisms. This paper and its companion \cite{Companion_Paper} introduce a synergistic, but {\em non-anthropomorphic}, robot hand. The companion paper showed how the mechanism's joint structure and link design allow this device to securely grab virtually all planar objects, while only using a single actuator.  It is thus synergistic, but not nearly as limited to a specific class of grasps. This paper presents design guidelines for this novel class of grippers, which possess a vast unexplored design space.

\begin{figure}[t]
    \centering
    \includegraphics[scale=0.6]{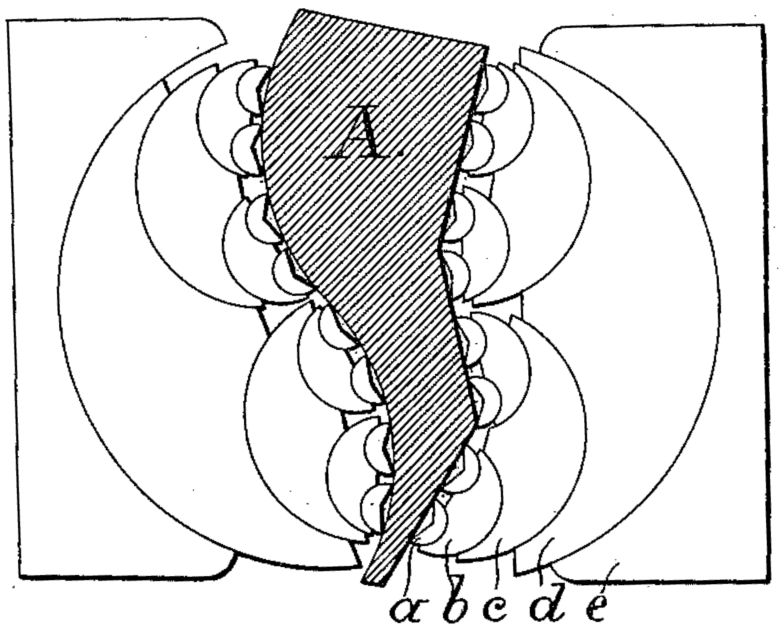}
    \vskip -0.1 true in
    \caption{An illustration from the origion ``Fractal Vise" patent document with, the fixtured object \(A\), fingers \(\alpha\), layers of connecting bodies \(b-e\), and translating body \(e\) \cite{VisePatent}}
    \label{fig:patent diagram}
    \vskip -0.2 true in
\end{figure}

In 1913, Paulin Karl Kunze patented a ``Device for obtaining intimate contact with, engaging, or clamping bodies of any shape \cite{VisePatent}." The device has only 1 actuated degree-of-freedom (the closing power screw of a traditional vise), but $2^{n+1}-1$ joints, where $n$ is a design parameter that defines the depth of the finger mechanism's binary tree structure. Fig. \ref{fig:patent diagram} shows a planar design, but Kunze's patent also presented 3-dimensional versions of the concept. The original design used sliding rotary dovetail joints to allow the mechanism ``cheeks" to rotate around a fixed point not coincident with the cheek itself (see  Fig. \ref{fig:patent diagram}). The joints could conceivably be actuated or compliant to provide asymmetries in fingertip reaction forces. Kunze noted possible uses including: clamping, surgical devices, supporting devices, and machinery equipment. This invention, later called the ``Fractal Vise" \cite{HandToolRescue, SavageFractalViseVideo}, primarily found a niche in machinists' shops. Although written in 1912, each of the patent's proposed uses describe areas of current robotic manipulation interest. While whippletree mechanisms share tree structures and load sharing capabilities with the Fractal Vise \cite{whippleHand,Tanaka2021AnUW, telescope} and are used in applications like load sharing between robotic fingers, cable car supports, rocker-bogie suspension systems, and under large optical telescope elements, these are only able to accommodate relatively small joint displacements. Unfortunately, manufacturing limitations available in Kunze's time led to poor functionality of the device, ultimately stalling his vision. 

This paper and its companion expand upon Kunze's work to propose a new class of {\em Fractal Hand} robot grippers. The companion paper describes the kinematics and properties of Kunze's mechanism \cite{Companion_Paper}. This paper focuses on broadening Kunze's vision. Section \ref{sec:descipt} describes the Fractal Hand and its relevant design parameters. Section \ref{sec:Problem Definitions} describes the design features and constraints needed to construct an effective Fractal Hand gripper. Section \ref{sec:Design Methodology} explains a method to synthesize a planar Fractal Hand. Section \ref{sec:grasps} highlights the planar Fractal Hands' capabilities via empirical tests. Section \ref{sec:Further Embodiments} presents preliminary ideas for other designs. 

%The conclusion anticipates the vast, unexplored design space yet to be navigated.

%we offer a broadened perspective on Fractal Hands, drawing inspiration from Kunze's original design.
% add figure 

%\section{The Vintage Device}
%\label{sec:The Fractal Vise}

%Given the attractive features described above and in the original patent, it is natural to ask why this device for "clamping bodies of any shape" never took hold. This choice was likely made based on the comparatively limited manufacturing capabilities of the time but makes the Fractal Vise susceptible to binding from manufacturing flaws and debris in the joints. While the vise was intended to change the paradigm of grasping in various fields, the stumbling blocks encountered in machining ultimately led to Paulin Karl Kunze's vision stalling. Now equipped with high precision, additive and subtractive, manufacturing techniques, we are more able to continue the legacy of the Fractal Vise. 

\section{Description of the Fractal Hand}
\label{sec:descipt}
%\note{This section should briefly describe the fractal hand. It can refer to the companion for detailed terminology and descriptive design parameters.  But need a bare minimum summary here. }

\begin{figure}[t]
    \centering
    \includegraphics[scale=0.4]{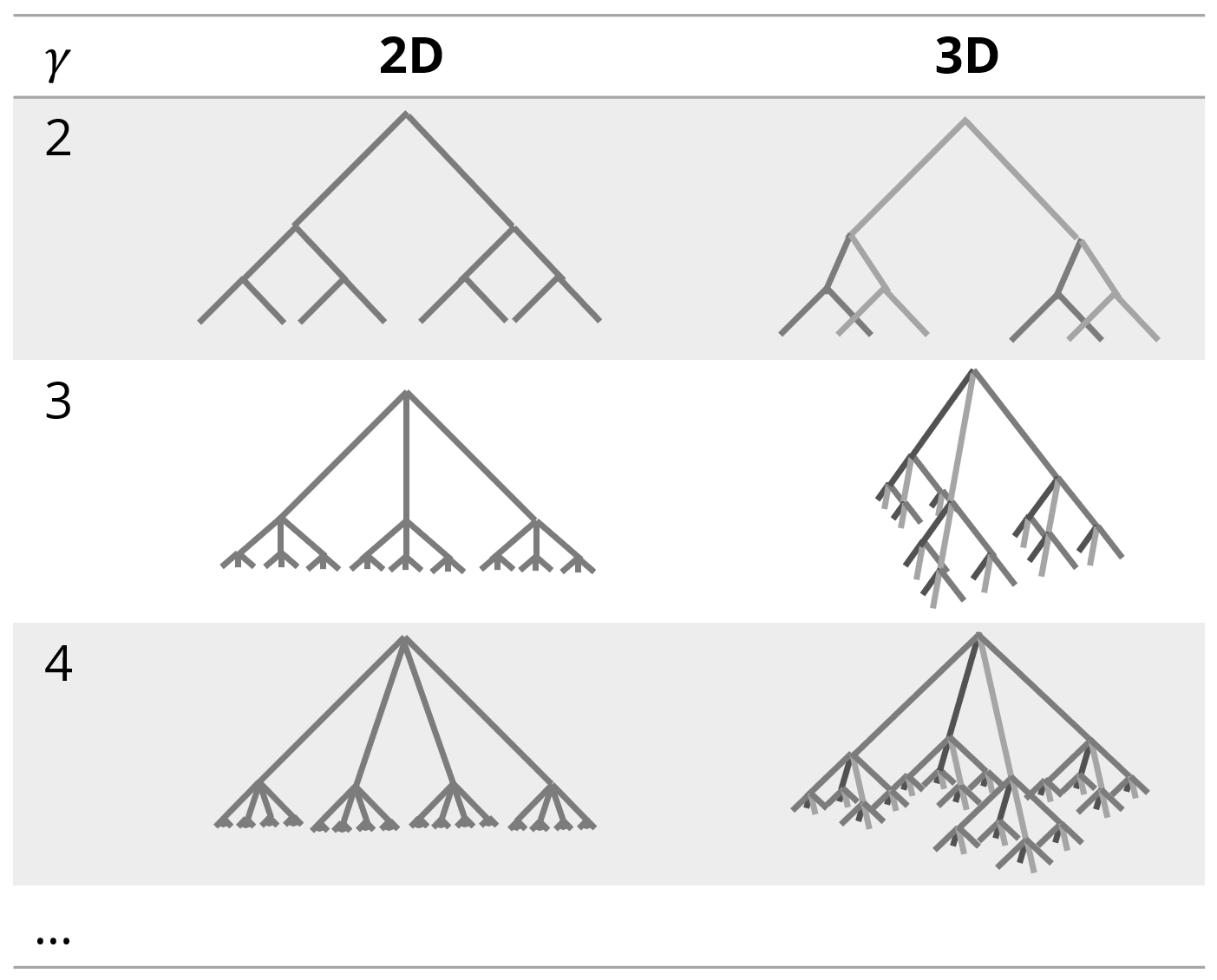}
    \vskip -0.1 true in
    \caption{Example two and three dimensional topologies with a uniform number, \(\gamma\), of children bodies per parent body and a joint tree depth of $n=3$.}
    \label{fig:topologies}
    \vskip -0.2 true in
\end{figure}

Kinematically, Fractal Fingers have a tree topology, where the nodes represent joints, and the edges represent rigid body links. All successor nodes and edges of a joint node will move with respect to that joint \cite{TreeSynthesis}. Let $\gamma_i=\frac{branches}{node}$ be defined at every node $i \in [1,...,N]$, where $N$ is the number of nodes. A tree is {\em uniform} if $\gamma_i = \text{constant } \forall i$, (see in Fig. \ref{fig:topologies}). Conceivably, any tree topology or combination of trees can be used as the basis of a Fractal Finger. The relevant design parameters for specifying a uniform, $\gamma=2$ Fractal Finger in this study are given below, and shown in Fig. \ref{fig:handschematic}: 
\begin{itemize}
    \item{} $n$ is the depth of the Fractal Finger joint tree.
    \item{} The {\em finger width}, $D$ is distance between farthest joints in the $n^{th}$ level.
    \item{} The {\em finger pitch}, $P$ is the distance between the $1^{st}$ and $n^{th}$ levels.
    \item{} The {\em joint stiffness constant}, $k$ relates joint displacements to joint torques.
\end{itemize}
Two or more Fractal Fingers move towards a common grasp center using a single actuator (see Fig. \ref{fig:handonwrist}). This study uses a prismatic actuator but other approaches are feasible. A more thorough description of parameters is given in the companion paper \cite{Companion_Paper}.

\section{Problem Definition}
\label{sec:Problem Definitions}

Inspired by Kunze's clamping device, we propose some general principles for fractal robotic fixturing fingers and hands that can adaptively acquire and hold an object rigidly with respect to the robot hand and arm. 
\begin{enumerate}
\item A Fractal Finger constitutes a tree topology of joints and connecting bodies that share loads between contact points and are conformable to complex surfaces,
\item Two or more Fractal Fingers combined with a closing actuator form a Fractal Hand (e.g., see Fig. \ref{fig:handonwrist}),
\item The placement of the finger links and joint axes should prevent internal mechanism collisions and afford a large workspace,
\item The fingertip's compliance (force reactions with respect to displacements) can be tailored by altering mechanism geometry or mechanical properties.
\end{enumerate}

\begin{figure}[t]
    \centering
    \includegraphics[scale=0.4]{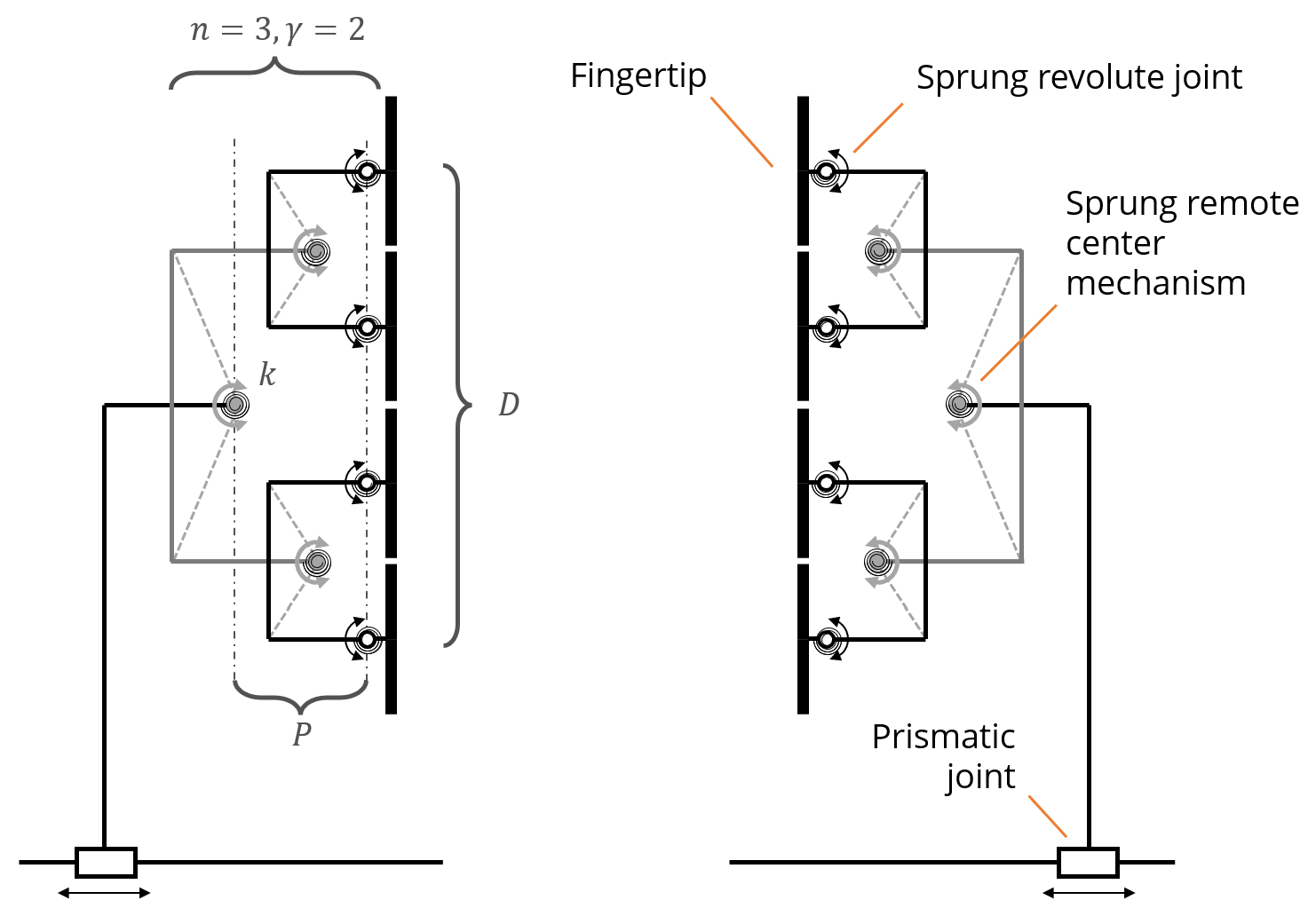}
    \vskip -0.1 true in
    \caption{Schematic of a Fractal Hand consisting of two 3-level Fractal Fingers coupled by a prismatic closing actuator. Relevant dimensions and components are labeled. As hand closes, the Fractal Fingers conform to an object's surface and provide fingertip contact forces.}
    \label{fig:handschematic}
\end{figure}

% In other words, a fractal vise embodiment is a whippletree mechanism that allows for large joint displacements due to each rotating joint having a remote center. 

%Trees are a specific type of graph consisting of nodes and edges which do not have any closed loops, like branches in a tree.  They exhibit the property that any two nodes or edges are only connected by one path.  Let $\gamma=\frac{branches}{node}$ defined at every node. We will consider a tree to be uniform if $\gamma$ is constant throughout the tree, as exemplified in Figure \ref{fig:topologies}. A non-uniform tree is shown in Figure \ref{fig:handtopologies}. Conceivably, any tree topology or combination of trees can be used as the basis of a Fractal Hand. Both whippletree mechanisms and the fractal vise share tree topologies where the the edges of tree are the rigid bodies and the nodes are the joints. If each node is a joint then all of the associated children, successor, nodes and edges will move with respect to that joint \cite{TreeSynthesis}. 

An $n$-level uniform Fractal Finger contains $\gamma^{n-1}-1$ joints, $\gamma^{n-1}$ fingertips, and a $\gamma^n$ scaling factor between joint bodies in the first and $n^{th}$-levels. Large $n$ and $\gamma$ heavily penalize design and manufacturing complexity. Thus, we aim to find the minimum $n$ and $\gamma$ to maintain universality within a expansive set of grasping tasks. We also aim to maximize the finger width $D$ while minimizing the pitch $P$ to yield a compact finger design. 

%A bounding area or volume can be constructed such that for all states of a given joint, all the successors remain inside that bounding box, as shown in Figure \ref{fig:rcmconflict}. To enable a Fractal Finger to conform to complex geometries, the mechanism must be designed such that theses bounding boxes never intersect, as this would result in self-collisions. The hierarchical nature of these mechanisms mean that bounding boxes must be contained completely within or outside of other bounding boxes. Maintaining this condition any number of children bodies can be associated with given parent joint, or node. A Remote Center Mechanism (RCM) at each joint allows rigid bodies around a point not contained within the mechanism, facilitating large displacements without collisions (a selection shown in Figure \ref{fig:rcmtrades}. In two dimensions and three dimensions, the bounding box takes the form an area or volume, respectively, centered at the Remote Center (RC). 

\begin{figure}[t]
    \centering
    \includegraphics[scale=0.18]{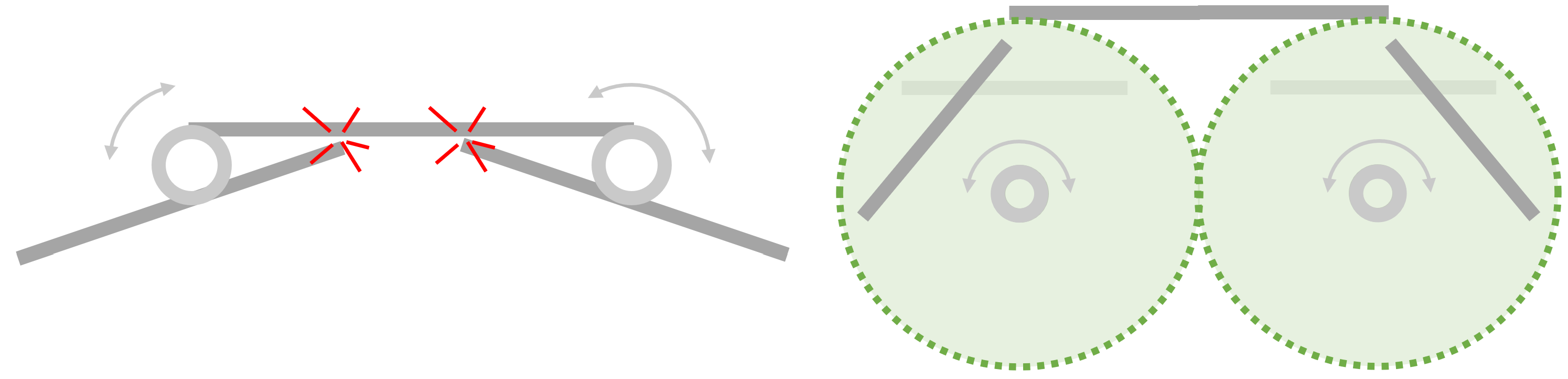}
    \vskip -0.1 true in
    \caption{The whipple tree mechanism on the left has a very small workspace.  Remote center mechanism (right) allow for large joint displacements while keeping the entire assembly planar and compact.}
    \label{fig:rcmconflict}
    \vskip -0.2 true in
\end{figure}

To prevent internal finger mechanism collisions, we note that a bounding area or volume relative to each joint in the hierarchy can be constructed such that for all states of the joint, its successors remain inside that bounding box (see Fig. \ref{fig:rcmconflict}). To accommodate complex grasped object geometries with a Fractal Finger and avoid self-collisions, these bounding areas must never overlap. Given the hierarchical mechanism structure, each bounding area must be entirely inside or outside others. Using a Remote Center Mechanism (RCM) at each joint, rigid bodies can maneuver around an external point, enabling significant displacements without internal interference. Thus, to generalize Kunze's concept, we focus on remote center mechanisms to allow design and manufacturing freedom of new Fractal Finger mechanisms.  
%In 2D and 3D, this bounding region centers around the Remote Center (RC) as an area or volume, respectively.

%If object contact points are equidistant from their respective centers of rotation and in equilibrium, equal magnitude contact forces will be generated. Uneven contact force profiles can be designed into the Fractal Hand with asymmetric contact locations or sprung joints. Tailoring the contact force profiles for differing object is an open subject of study.  

To showcase the \textit{Fractal Hand}'s potential grasping capabilities, we constructed a prototype that improves upon the issues found in the vintage design.

% \begin{figure}[t]
%    \centering
%    \includegraphics[scale=0.2]{images/handtopologies.png}
%    \caption{Diagram of a potential Fractal Hand tree structure. A Fractal Hand may have a fixed number of rigid bodies per joint, as in the Fractal Vise, or a varying number of rigid bodies per joint, as shown in the diagram. The tree may also have more and less dense regions}
%    \label{fig:handtopologies}
%\end{figure}

\section{Design Methodology}
\label{sec:Design Methodology}

%Designing a Fractal Hand embodiment raises a paradox. How can we synthesise a gripper with an idea of the objects it will grasp without constraining its universality? Specifically, the tree topology, corresponding depth, and arrangement in two or three dimensions must be chosen before more detailed design decisions can be made. 

To create an effective robotic gripper, the following design desiderata were observed to constrain the design space.
\begin{enumerate}
\item Maximize ability to conform to varied object geometries
\item Minimize electromechanical and manufacturing complexity
\item Minimize volume
\item Robustify against deformations 
\item Minimize grasping cycle time
\end{enumerate}

%This section addresses the kinematic synthesis problem of designing a Fractal Hand and is subdivided into tree synthesis, mechanism synthesis, and part synthesis. Tree synthesis tackles the problem of selecting the appropriate taskspace, tree topology, overall dimension, and corresponding depth. Mechanism synthesis is the process of deciding the combination of joints and links to satisfy the desired tree topology and their relevant dimensions. Part synthesis determines the design of the joints, links, and additional features to allow the Fractal Hand to function. 

The following subsection addresses the kinematic synthesis problem of designing a Fractal Hand.

%and is subdivided into tree synthesis, mechanism synthesis, and part synthesis. 

%Given the underlying fractal nature of the gripper, we created a tool that analyses the scale-variant complexity of object data sets to extract the necessary design parameters of a Fractal Hand.% tie this into layers and tree and location. Then complexity.....

\subsection{Tree Synthesis}
\label{subsec:mfd}

While a Fractal Hand could be applied to countless grasping tasks, we introduce below some basic principles to select the design parameters $\gamma, P, D, n$ for a given class of objects. We chose to initially focus on planar hands before exploring a spatial embodiment (discussed in Section \ref{sec:Further Embodiments}). 

In the context of the above design characteristics, a Fractal Finger with a uniform tree simplifies the design process. With uniformity, joints and links can be designed once and scaled or repeated throughout the mechanism. For a planar hand with 1-DOF joints, the choice of $\gamma=2$ (a binary tree) ensures moment balancing around each joint, without over constraining the mechanism. 

To minimize the footprint of a finger, we set $P=0$, meaning each remote center is arranged alone a common line at the front surface of the gripper.

The Fractal Hand adapts to a wide variety of shapes.  But clearly, its overall dimension should be compatible with the range of object sizes that are expect to be grasped. To choose the gripper width and its tree depth, we examine several factors.  

Assuming that each joint has unlimited rotation, a reasonable upper bound for $D$ is the perimeter of the object divided by the number of Fractal Fingers, allowing the object to be enveloped. 

%To calculate $D$, imagine the limit as $n \rightarrow\ \infty$.
%For large $n$, the gripper's contacting surface can conceptually be viewed as a rope that conforms to and provides contact forces along the object perimeter. 
%It is evident that this would be an equilibrium, wrench closure grasp, given sufficient normal force. However, wrench closure could not be achieved with frictionless grasps on the exceptional objects of a circle and an infinitely long bar. 
% With this restriction, we can examine the relationship between wrench closure and $n$. Wrench closure is attained when the grasp map, with an appropriate friction model, positively spans the wrench space. When more than one finger shares equivalent surface normals they are linearly dependent and thus redundant. Adjacent fingers may have the same surface normals when the surface is locally linear. Moment closure is achieved when the gripper contacts at points are in a neighborhood of a point where the change in curvature changes signs.

\begin{figure}[t]
   \centering
    \includegraphics[scale=0.23]{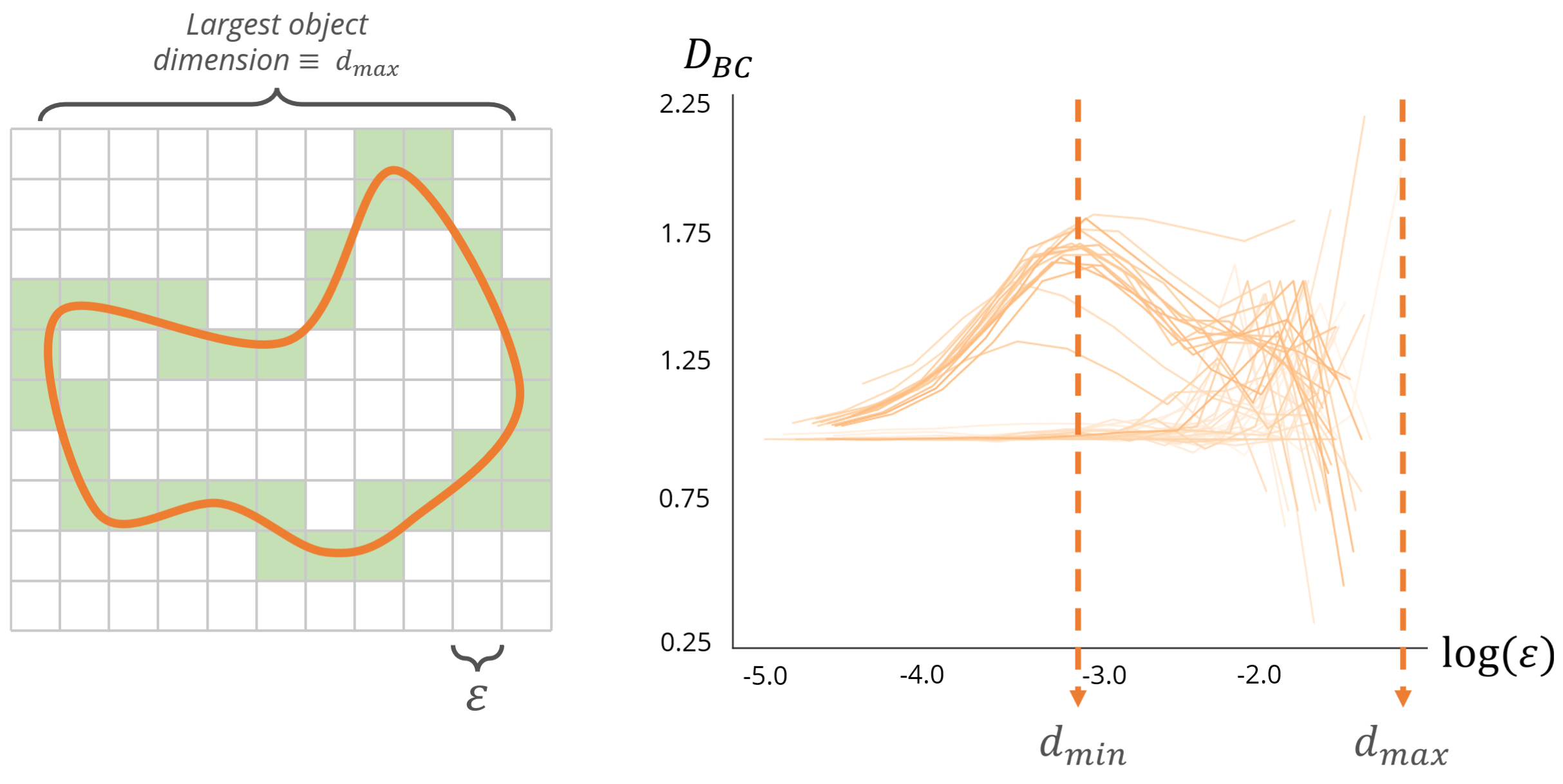}
    \caption{Left: Illustration of box counting method where the green boxes are totaled for a given grid-size, $\epsilon$. Right: Plot of Box Counting Dimension, $D_{BC}$ vs. $\epsilon$, based on the 2D cross sections of 50 household objects. Smaller $\epsilon$ converges to a dimension of 1}.
    \label{fig:fracdimeps}
    \vskip -0.2 true in
\end{figure}

\begin{figure*}[t]
    \centering
    \includegraphics[scale=0.55]{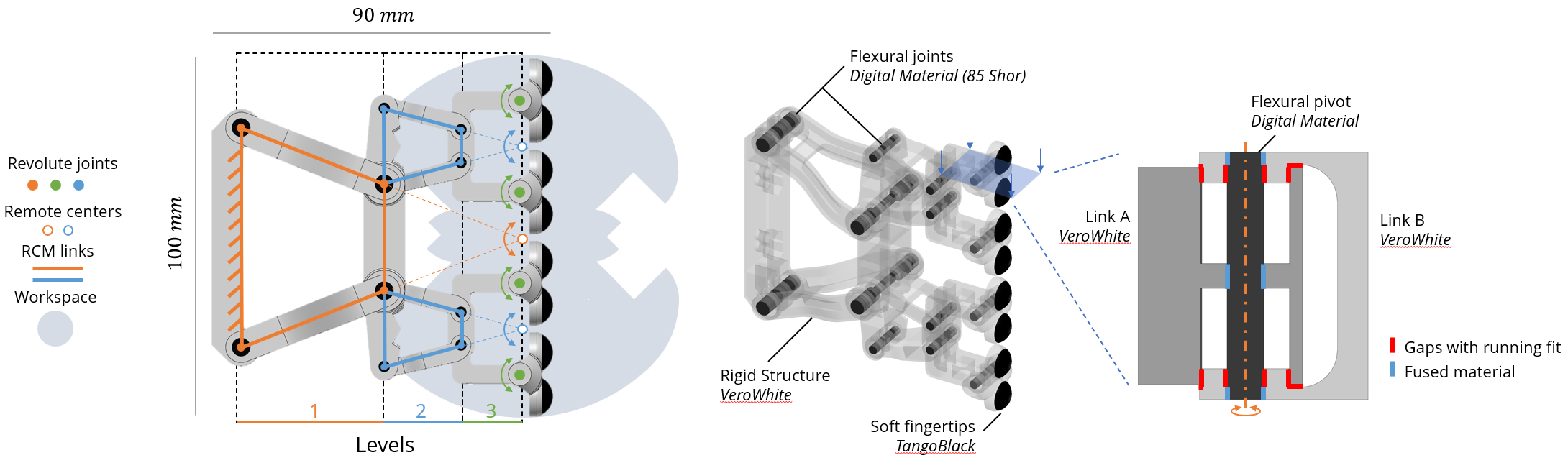}
    \vskip -0.1 true in
    \caption{Left: Diagram of a 3-level, Fractal Finger module. Each level, colored, rotates around corresponding remote centers. Levels 1 and 2 use an isosceles trapezoidal RCM, while level 3 employs revolute joints. Center: Each Fractal Hand module is 3-D printed simultaneously with rigid structural material and flexible compliant components. The joints consist of a digital material which is a mixture of rigid and flexible material for a desired stiffness. The rigid material is made translucent for clarity. Right: Cross section view of the flexural joint design from the blue cutting plane. Rigid materials, shown in shades of gray, rotate around a cylinderical region of flexible material. When torsionally loaded, the flexible material acts as a spring. Gaps with a running fit, shown in red, provide bushing-like off axis rigidity. The blue lines show where differing materials were fused by the 3D printing process.}
    \vskip -0.15 true in
    \label{fig:mechdiagram}
\end{figure*}

Next, we consider how each Fractal Finger can locally conform along its surface length.  The spacing between fingers should be sized to capture the characteristic changes in object curvature. This design factor can be approximated using the following construction. We use the {\em Box Counting Dimension} $D_{BC}$, defined in Eq. (\ref{eqn:mink}), to estimate the complexity of an object with respect to a grid size, $\epsilon$ \cite{fractalgeometry}. The box counting method covers an object with grid having spacing $\epsilon$. The number of boxes $N$ intercepted by the object boundary is found (see Fig. \ref{fig:fracdimeps}). As $\epsilon \rightarrow \infty$, $D_{BC}$ converges to the Upper Minkowski dimension. 
\begin{equation}
   \label{eqn:mink}
   D_{BC}(\epsilon) = {\frac{-log(N(\epsilon))}{log(\epsilon)}}
\end{equation}

Although the box counting dimension is not a unique descriptor of individual objects, it can describe an object's complexity with respect to the characteristic size $\epsilon$. When the box counting dimension converges to one, the object is locally straight when examined at scale $\epsilon$. We propose that in order to match the complexity of a set of objects, a Fractal Hand must accommodate the largest object size, $d_{max}$ and most "complex" scale, $d_{min}$, past which the complexity of the objects converge. A sufficient upper bound for the $d_{max}$ can be evaluated from $\operatorname{max} (\text{object length})$. Then the number of layers can be approximated in Eq. (\ref{eqn:n}).
\begin{equation}
   \label{eqn:n}
   n = \lceil \frac{log(\frac{d_{max}}{d_{min}})}{log(\gamma)} + 1 \rceil
\end{equation}

%\note{I don't understand the phrase " the algorithm was checked for convergence at small $\epsilon$ against objects with known dimensions"}

%\note{The values of $d_{max}$ and $n$ are not updated at the end of the next paragraph.}

%. To minimize the effect of quantization errors, the algorithm was checked for convergence at small $\epsilon$ against objects with known dimensions

The $D_{BC}$ was evaluated for random cross sections of 50 household objects, including: boxes, tools, fruit, and cups \cite{li2023frogger}, resulting in the plot of Fig. \ref{fig:fracdimeps}. The $D_{BC}$ dimension tends towards 1 as $\epsilon \rightarrow 0$, which represents line-like features. A bimodal behavior is observed as some objects converge immediately (from right to left) while others peak around $log(\epsilon) = -3.25$ before approaching 1, likely due to surface irregularities. Aside from the $D_{BC}$, the plot gives a sense of the objects' complexity with respect to scale. Using Eq. (\ref{eqn:n}), a Fractal Hand for the objects in this sample set should have $d_{max}=D=30\text{cm}$, and at most $n=9$. This method gives a conservative upper bound for the number of levels, which will likely be limited by manufacturing capabilities. 

Our prototypes have $D=100\text{mm}$ to provide a compact gripper size while having sufficient footprint to fit 3 levels ($n=3$). A hand built from fingers of this size can conform to a wide array of objects, without requiring special manufacturing processes for the miniaturization of the final levels of the mechanism.

\subsection{Mechanism Synthesis} 
\label{subsec:mechsynth}
The original fractal vise's sliding dovetail joints are difficult to manufacture, and easily malfunction in the presence of dust and debris. Moreover, this design necessarily results in large, bulky, and heavy Fractal Fingers. This section describes our process of arriving at a new Fractal Finger design based on a remote center of compliance mechanism. This design is lighter in weight, readily manufactured using a single-step 3-D printing process, and has a large workspace.

\subsubsection{Remote Center Mechanism Design}
\label{subsubsec:rcm}

%\begin{figure}[t]
  %  \centering
    %\includegraphics[scale=0.3]{images/rcmtrades.png}
  %  \caption{The five shown remote center mechanisms were downselected from literature review for further trade studying. a) Semicircular slider, b) Parallelagram, c) Isosceles trapezoidal, d) Ring pantograph, e) Duel Triangular. Remote centers shown in orange. Selected mechanism highlighted in green}
   % \label{fig:rcmtrades}
%\end{figure}

% RCM type and dimensions
%After choosing the Fractal Hand topology and overall design parameters, the specific mechanisms and dimensions to achieve the properties stated in Sections \ref{sec:Problem Definitions} and \ref{sec:Design Methodology} must be selected. 
Kunze's fractal vise is based on a tree topology of revolute joints.  We sought to replace the revolute joint with an RCM which can also more readily integrate tunable compliance.  Various RCM types were considered, including sliding, parallelogram, ring pantograph, dual triangular, and isosceles trapezoidal linkages. The isosceles trapezoidal linkage, composed solely of revolute joints, allows for adjustable torsional stiffness through joint modulation and offers favorable scaling with its four similarly sized bars. The Remote Center (RC), $O$, remains stationary for small angular perturbations of the RCM from equilibrium. However, to accommodate large angular displacements, a dimensional synthesis method optimized the relative link lengths that minimize the RC drift, $\delta$. 

%\note{The next paragraph needs to be shortened. Do we really need to discuss bounding box?  Can we just summarize the optimization.  }

%we constructed a circular bounding box centered on $O$, and a reduced bounding box centered on $O'$, tangent to $O$. Given a maximum desired angular deflection of each RCM, 

\begin{figure}[t]
    \centering
    \includegraphics[scale=0.33]{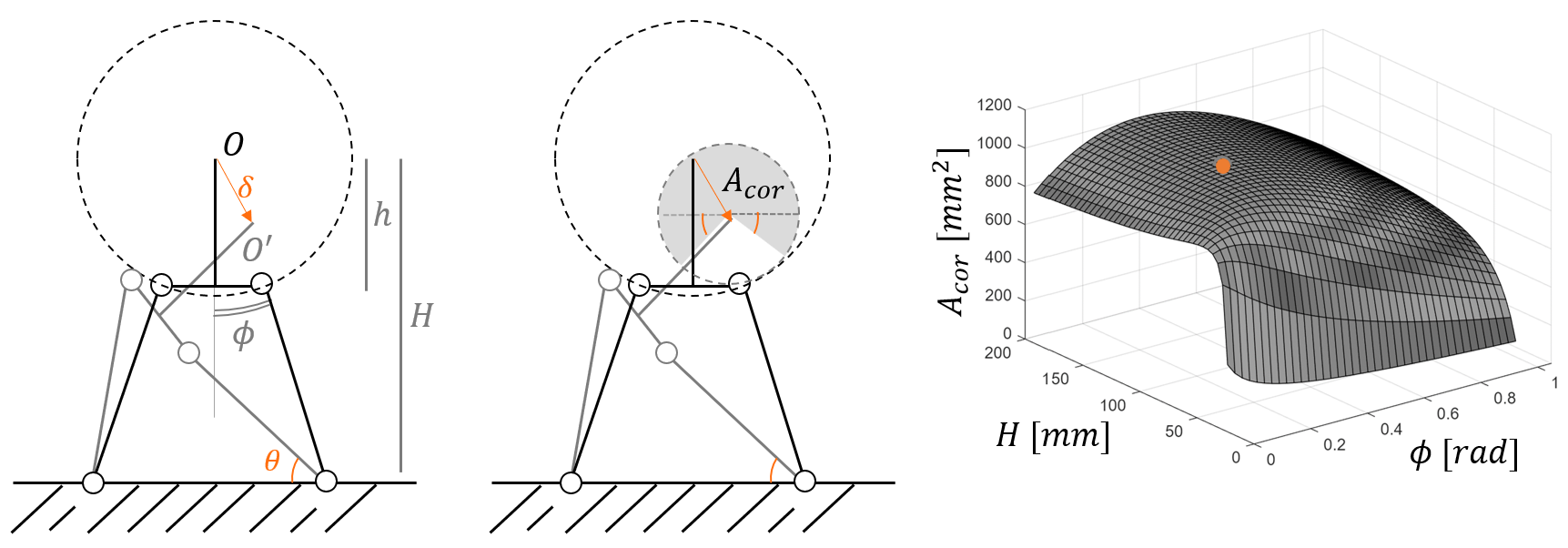}
    \caption{Left: The isosceles trapezoidal linkage is defined by $\phi$, $h$, and $H$, and maximum allowable angle $\theta$. $O$ drifts, $\delta$ to $O'$. Right: The design metric $A_{cor}$ vs. key trapezoid parameters. The local maxima is highlighted by an orange circle}.
    \label{fig:dimensionalsynth}
    \vskip -0.2 true in
\end{figure}

Each isosceles trapezoidal RCM is defined by a bar angle $\phi$, mechanism height $H$, and clearance height $h$, see Fig. \ref{fig:dimensionalsynth}. To avoid self-collisions due to RC drift and maximize workspace, we would like to minimize $\delta$ to maximize the corrected area of the bounding box on $O'$, $A_{cor}$. Using the kinematics formulated by Xu et al, $A_{cor}$ was evaluated for a fixed $h$ and variable $\phi$ and $H$ \cite{iscostraplinkage}, as shown in Eq. (\ref{eqn:rcmcost}). 

\begin{equation}
   \label{eqn:rcmcost}
   A_{cor}=\theta(\phi,h,H)(h-\delta(\phi, h, H))^2
\end{equation}

An example design space is shown in Fig. \ref{fig:dimensionalsynth}, with the maxima highlighted in orange. Sequentially, the dimensions of each RCM can be defined by the level below it, with the next level's $h$ equal to the previous level's $H$. At the $n^{th}$ level, $h=\frac{D}{2\gamma^{n-1}}$. The resulting trapezoidal link lengths and RCs are shown in Fig. \ref{fig:mechdiagram}, superimposed on the 3D model. 

%\begin{figure}[t]
 %   \vskip -0.15 true in
  %  \centering
   % \includegraphics[scale=0.35]{images/linkageplot.png}
    %\vskip -0.1 true in
    %\caption{The design metric $A_{cor}$ vs. key trapezoid parameters. The local maxima is highlighted by an orange circle}
    %\label{fig:linkageplot}
    %\vskip -0.17 true in
%\end{figure}

%\begin{figure}[t]
   % \centering
    %\includegraphics[scale=0.3]{images/jointtrades.png}
   % \vskip -0.1 true in
   % \caption{The five shown flexural pivots were downselected from literature review for further studying and prototyping. a) notch, b) cartwheel, c) leaf-spring, d) cross torsional, e) circular torsional. Flexible material shown in black, rigid material in white. The selected design is highlighted in green}
   % \label{fig:jointtrades}
   % \vskip -0.15 true in
%\end{figure}

\subsubsection{Parallel Actuator Design}
\label{subsubsec:actuator}

The Fractal Hand only requires one active closing actuator to provide stable grasps with $\gamma^{n-1}-1$ joints. 
%To prevent unnecessary sliding between fingertips and a given object, the hand should translate along a line that is perpendicular to the line or plane shared by the fingertips. 
The two fingers can be relatively oriented in a variety of ways.  We chose to place the fingers in direct opposition, since this configuration mimics standard parallel jaw grippers.  
%Each opposing hand must have a stiff to the motion of the other such that once a hand has fully conformed to an object, no translation is allowed that could potentially let an object be displaced. 
The stroke length of the closing actuator can be chosen arbitrarily depending on objects' size. To capitalize on the load sharing capabilities of Fractal Hand fingertips, a passive center 1-DOF prismatic drive system was constructed as shown in Fig. \ref{fig:handonwrist}. 

\subsection{Part Synthesis}
\label{subsec:partsyn}

\subsubsection{Flexural Pivots}
\label{subsubsec:flexpivots}

Incorporating spring-like behavior into the RCM joints has several potential benefits. Such springs automatically return the hand to a neutral pose upon object release. Tuned springs can also endow a hand with specific fingertip compliance response functions.

To enable trapezoidal linkages to rotate with a prescribed stiffness, and to minimize out of plane bending, we elected to use flexure based joint springs. Using a multimedia 3D printer (Stratsys Objet 350), flexible (TangoBlack mixed with VeroWhite) and rigid materials (VeroWhite) can be printed together in complex geometries to build the entire Fractal Hand in one print. Several flexural joint types were studied. A circular torsional joint type was chosen as it provides a stationary axis of rotation, easily tunable stiffness, and minimal out of plane bending. A cross section view is shown in Fig. \ref{fig:mechdiagram}. Rigid joint features act as flanged bushings on either side of the flexural pivot, increasing off axis stiffness. The joint's torsional stiffness can be simply tuned by increasing the material stiffness or diameter of the cylindrical flexible material. While higher joint stiffness leads to faster returns of a gripper to its rest position, it also lead to greater asymmetries in fingertip contact forces when the finger joints are displaced.
% The affect of spring stiffness on grasp stability is discussed in the companion paper. 
Our planar Fractal hand prototype used a 85 Shor digital material. 

\subsubsection{Fingertips}
\label{subsubsec:fingertips}

To allow the fingertips to conform to surfaces of varying curvature, two rubber hemispheres (soft TangoBlack material) are printed on either side of the fingertip. They provide additional friction and contact compliance. Other fingertip concepts are currently under study.

\subsubsection{Integrated Structure}
\label{subsubsec:structure}

Fig. \ref{fig:mechdiagram} shows a planar Fractal Hand prototype, with its materials and kinematic relations denoted. Precise 3D printing enables design features that let joints pass through one another to maximize the gripper's conformability. Each RCM can rotate $\pm 45^{\circ}$ until it reaches a hard-stop that prevents excessive joint deformation. The final level can rotate through $\pm 90^{\circ}$. The first and second level pivots are coaxial to improve mechanism compactness. The final dimension of one planar hand are $100 \text{mm}$ by $90 \text{mm}$ by $20 \text{mm}$, at $37 \text{g}$. Ribs and beams add mechanism rigidity. 

\section{Fractal Hand Grasping}
\label{sec:grasps}

\subsection{Demonstrations}
\label{subsec:testing}

\begin{figure}[t]
    \centering
    \includegraphics[scale=0.16]{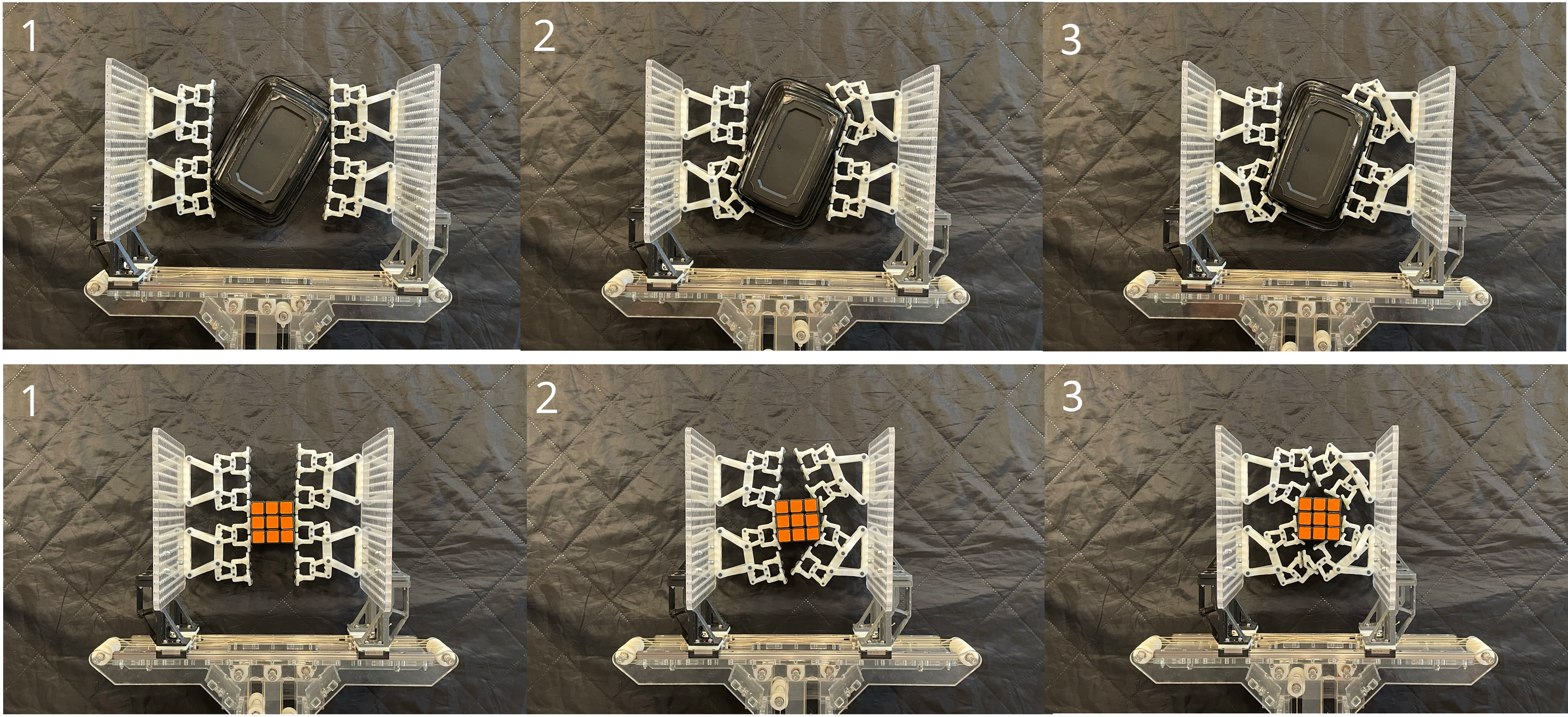}
    \caption{Photos (1-3) of opposing Fractal Fingers closing around two objects. Top: grasping an object larger than the hand span.  B: grasping an object smaller than the hand span. When the hand grasps a smaller object, opposing fingertips meet.}
    \label{fig:graspingshots}
    \vskip -0.3 true in
\end{figure}

\begin{figure*}[t]
    \centering
    \includegraphics[scale=0.44]{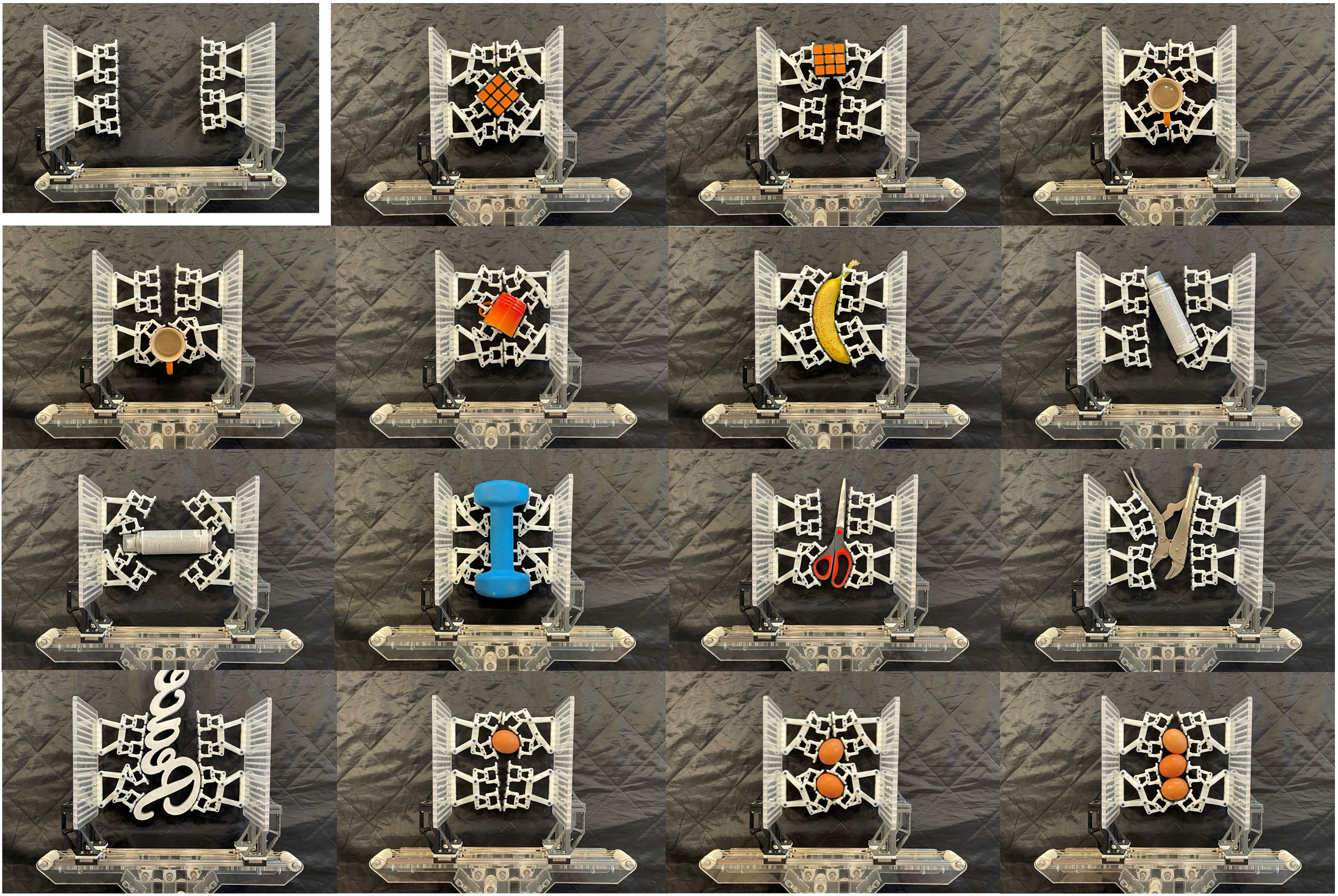}
    \caption{Grasps of 15 objects in increasing complexity.}
    \label{fig:grasps}
    \vskip -0.2 true in
\end{figure*}

To demonstrate our planar Fractal Hand design's ability to conform to a variety of object geometries, the gripper was mounted on a parallel jaw actuator and used to grasp objects of varying complexity, rigidity, and scale. The gripper, centered on the object, is closed by shortening the tendons in the passive center prismatic actuator using a hand crank instead of a rotary motor. The process of closing around an object both smaller and larger than the gripper is shown in Fig. \ref{fig:graspingshots}. When the gripper span is larger than the object, opposing fingers meet and immobilize each other. Fig. \ref{fig:grasps} shows the Fractal Hand grasping 15 objects of sizes and profiles. With four Fractal Fingers present, only two are needed to achieve an equilibrium grasp, so some might not contact the object.

\subsection{Comparisons with Antipodal Point Gripper}
\label{subsec:comparison}
\begin{figure}[t]
    \centering
    \includegraphics[scale=0.22]{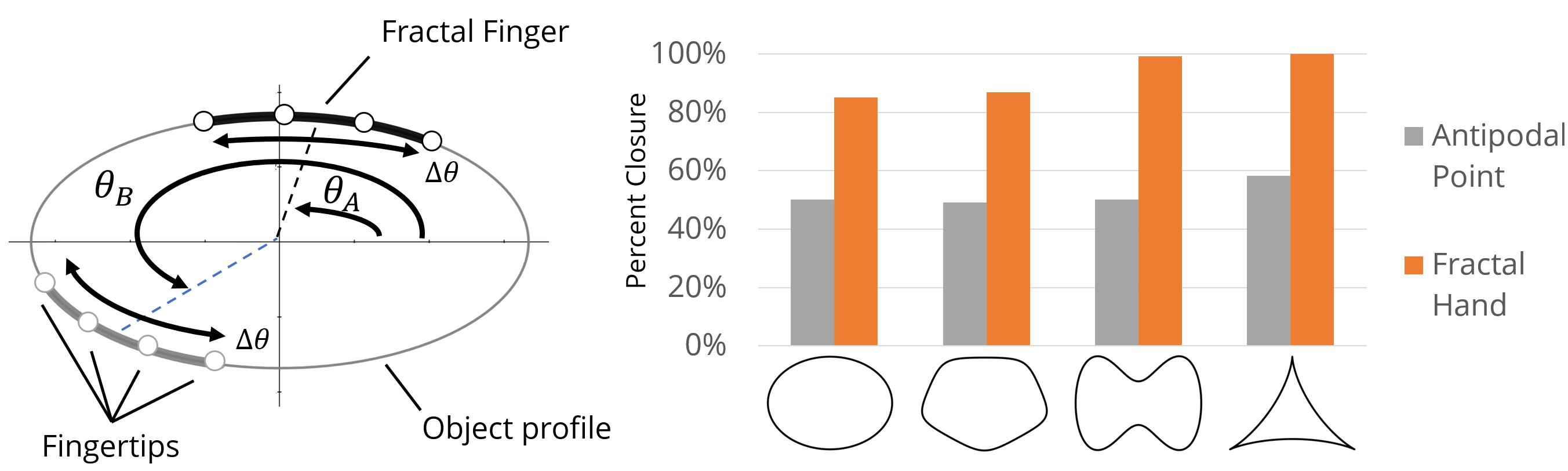}
    \caption{Left: An example 2D object profile in gray on the left. The thick segments signify the Fractal Fingers each defined by $\theta$ with $\Delta\theta$. Circles correspond with fingertip position if $n=3$.  Right: Plot showing percentage of configuration space leading to wrench closure using the antipodal point grasp (gray) and the Fractal Hand (orange), for each object profile.}
    \label{fig:ellcspaceFH}
    \vskip -0.2 true in
\end{figure}

\vskip -0.05 true in
Two finger parallel jaw grippers are widely used, and serve as an appropriate point of comparison for the Fractal Hand since both use a single actuator. % An existing parallel jaw gripper could be outfitted with Fractal fingertips. 
To compare their ability to achieve secure grasps, we studied the configuration space of Fractal Hand and antipodal point grasps on an elliptical, smoothed pentagonal, triangular, and dogbone shapes (see Fig. \ref{fig:ellcspaceFH}). We modeled the The Fractal Hand as a set of $2^{n-1}$ frictional point contacts spread along a prescribed grasp length $D$ coincident with the object boundary, assuming the fingertips will be evenly spread across the grasp length. For the four objects being studied, $D=\Delta\theta=2.7 \text{rad}$ and $n=5$, per the method described in Section \ref{subsec:mfd}, assuming arc length is proportional to $\Delta\theta$. The configuration space of grasps on these planar curves then consists of the center positions of each hand mapped onto the parametric curve parameter $(\theta_A, \theta_B) \in [0, 2\pi]$ in increments of $\frac{2\pi}{99}$ (shown in Fig. \ref{fig:ellcspaceFH}). The antipodal frictional point grasp configuration space was defined similarly \cite{Grasp_Book}. The Fractal Hand was able to achieve wrench closure an average of $40.9\%$ more than the antipodal point grasps over the entire configuration space across the four objects. See Fig. \ref{fig:ellcspaceFH} for the performance comparison in each object. As expected, the Fractal Hand outperforms the parallel jaw gripper in every case. The percent coverage of the configuration space was found to increase with $D$ and $n$. 

\section{Further Embodiments}
\label{sec:Further Embodiments}

\begin{figure}[t]
    \centering
    \includegraphics[scale=0.1]{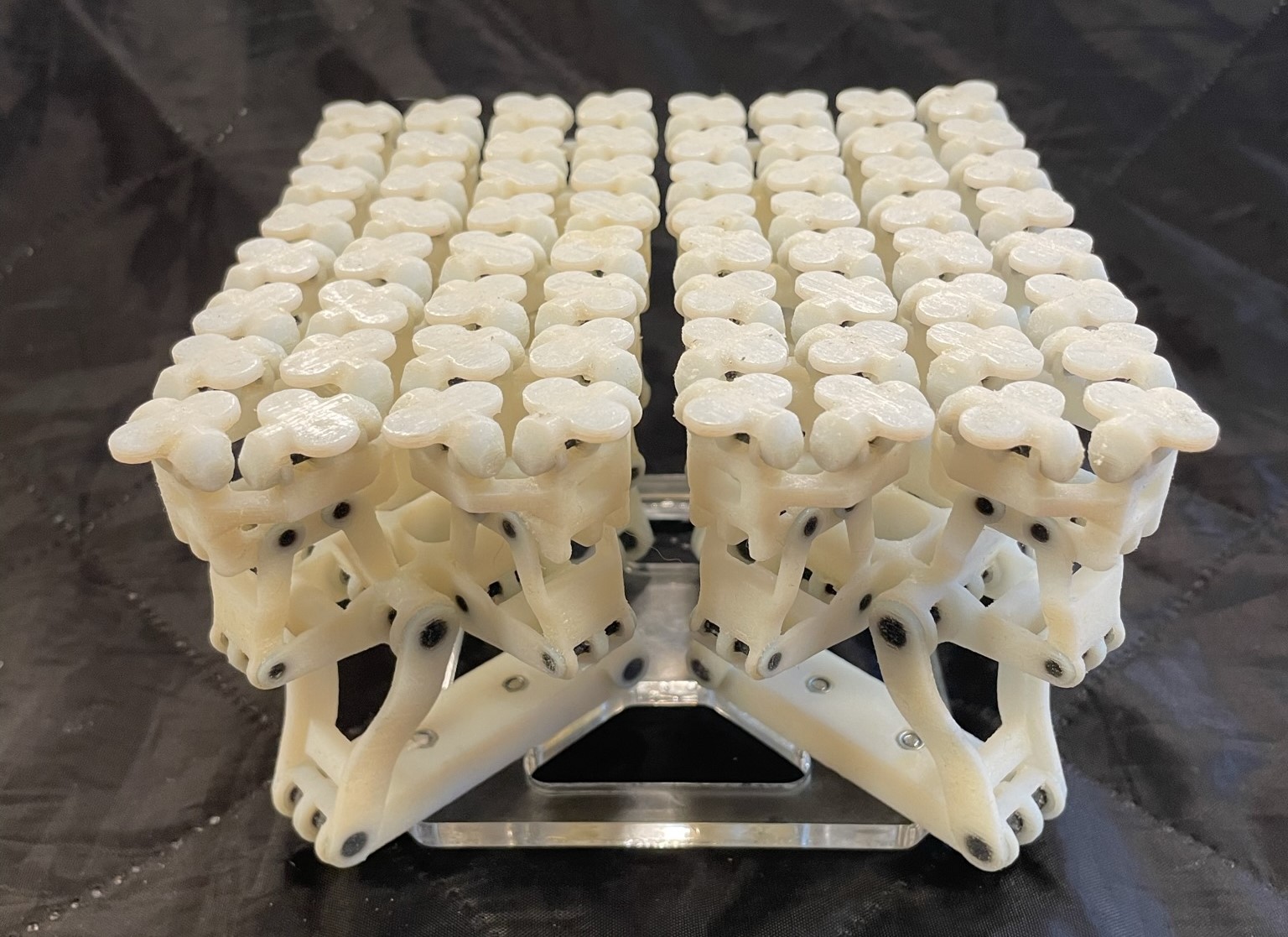}
    \vskip -0.05 true in
    \caption{Photo of four sets of 3-D Fractal Fingers.}
    \label{fig:3dhand}
    \vskip -0.2 true in
\end{figure}

While our prototype Fractal Hand performs in complex planar robotic grasping tasks, the Fractal Hand concept can be extended to 3-dimensional grasps. A $\gamma=2, n=3$ spatial Fractal Finger based on 1-DOF joints is shown in Fig. \ref{fig:3dhand}. With 1-DOF joints, this gripper can only conform to roughly spherical objects. However, with 2-DOF joints and $\gamma=3$ we envision a hand that can conform to complex spatial surfaces. Such a device can also serve as an adaptive foot.

% While the Fractal Hand presented above utilizes mechanical couplings to balance contact forces and couple fingertips, other mechanisms could achieve a similar effect. Couplings could be generated fluidically, electromechanically, or thermally. Within mechanical couplings, the prismatic joints could replace rotating joints. 
The driving motivation for the Fractal Hand is to eliminate mechanical, actuation, and sensing complexity, while still enabling high adaptivity. 
However, joint sensing, fingertip force sensing, and selected joint actuation, could provide additional real-world benefits. 
 
% To validate the embodied intelligence of the fractal hand, the success rate of randomized planar grasps was evaluated against objects from the fractal dimension data set. A 3D printed embodiment of the hand was mounted on a Franka Emika 7-DoF arm. The opposing sides of the fractal hand were mounted on a Franka Emika FE parallel jaw gripper. Two ZED 2i stereo cameras were positioned around arm workspace for object detection and point cloud generation. Adversarial objects......% figure showing pictures of different grasps especially with adversarial objects 

\section{Conclusions}
\label{sec:Conclusions}
The Fractal Hand is a promising robotic gripper in its ability to fixture objects of varying geometry, stiffnesses, and pose using only one actuator. Moreover, it shares the benefits of compliance in soft hands while having well described grasp stiffness and kinematic properties as shown in the companion paper \cite{Companion_Paper}. In this way, the gripper can be simply added to existing parallel jaw grippers to simplify the requirements of perception and grasp planning pipelines. The kinematic synthesis method proposed provides necessary bounds for the design of a uniform binary tree Fractal Finger. Future work will investigate sufficient bounds to the finger width $D$, and depth $n$, that could better allow the Fractal Hand to operate in confined spaces. Subsequent investigations will delve deeper into the extensive design landscape of Fractal Hands, examining their potential in pick and place tasks, adaptive feet, spacecraft manipulation, underwater sample collection, and medical robotics, among other applications.

%Discuss need to find sufficient lower bounds of levels and size especially in the need to operate in confined spaces

%Discuss potential to simplify grasp planning

%low cost improvement to existing parallel jaw grippers 

%more testing needed 

\bibliography{root}
\bibliographystyle{IEEEtran}

\end{document}